\documentclass[conference]{IEEEtran}
\usepackage{cite}

\ifCLASSINFOpdf
   \usepackage[pdftex]{graphicx}
  \graphicspath{{./fig/}}
   \DeclareGraphicsExtensions{.pdf,.jpeg,.png,.jpg}
\else
\fi
\usepackage[cmex10]{amsmath}
\usepackage{array}

\usepackage{algorithm2e}
\usepackage{algpseudocode}

\newcolumntype{V}{>{\centering\arraybackslash} m{.4\linewidth} }

\usepackage{amsmath}

\begin{document}
%
\title{Efficient Scene Text Localization and Recognition with Local Character Refinement \vspace{-10pt}}

\author{\IEEEauthorblockN{Luk\'{a}\v{s} Neumann}
\IEEEauthorblockA{Centre for Machine Perception, Department of Cybernetics\\
Czech Technical University, Prague, Czech Republic\\
neumalu1@cmp.felk.cvut.cz}
\and
\IEEEauthorblockN{Ji\v{r}\'{\i} Matas}
\IEEEauthorblockA{Centre for Machine Perception, Department of Cybernetics\\
Czech Technical University, Prague, Czech Republic\\
matas@cmp.felk.cvut.cz}}

\vspace{-15pt}

\maketitle

\begin{abstract}
An unconstrained end-to-end text localization and recognition method is presented. The method detects initial text hypothesis in a single pass by an efficient region-based method and subsequently refines the text hypothesis using a more robust local text model, which deviates from the common assumption of region-based methods that all characters are detected as connected components.

Additionally, a novel feature based on character stroke area estimation is introduced. The feature is efficiently computed from a region distance map, it is invariant to scaling and rotations and allows to efficiently detect text regions regardless of what portion of text they capture.

The method runs in real time and achieves state-of-the-art text localization and recognition results on the ICDAR 2013 Robust Reading dataset.
\vspace{-10pt}
\end{abstract}


%
\IEEEpeerreviewmaketitle

\section{Introduction}
Scene text localization and recognition, also known as text-in-the-wild or PhotoOCR, is an interesting problem with many application areas such as translation, assistance to the visually impaired or searching large image databases (e.g. Flickr or Google Images) by their textual content. But unlike traditional document OCR, none of the scene text recognition methods has yet achieved sufficient accuracy and speed for practical applications.

Text localization can be computationally very expensive because in an image of $N$ pixels in general up to any of the $2^N$ subsets can correspond to text. Methods based on the sliding-window localize individual characters~\cite{Wang-ICCV2011,photoocr} or whole words~\cite{SlidingWindow-ICDAR11} by shifting a classification window of multiple sizes across the image, drawing inspiration from other object detection problems where this approach has been has been successfully applied~\cite{viola2004robust}. Such methods are robust to noise and blur, since features aggregated over a larger area, but the crucial disadvantage is that the number of windows that needs to be classified grows rapidly when text with different scale, aspect, rotation and other distortions has to be found.

\begin{figure}
\centering
\begin{tabular}{VV}
\includegraphics[height=2.3cm]{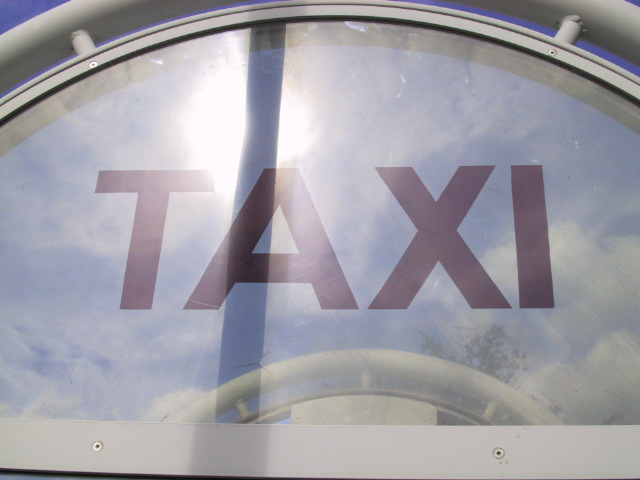}& \includegraphics[height=2.3cm]{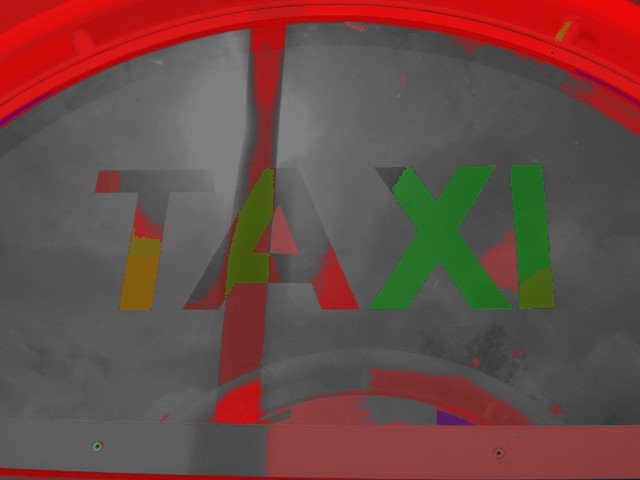}  \\
\small(a) &\small (b) \\
\includegraphics[height=2.3cm]{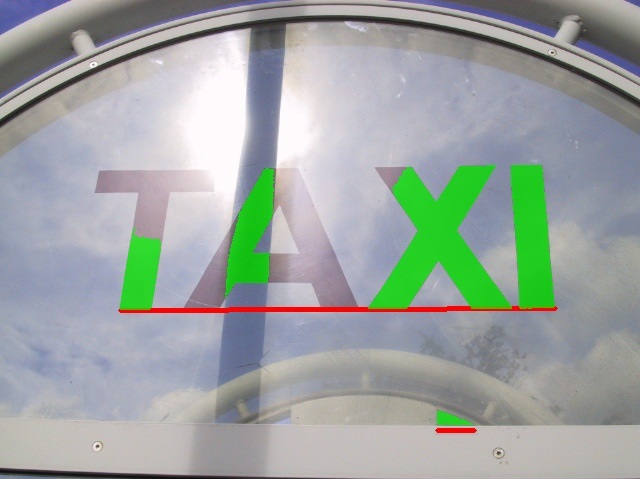} & \includegraphics[height=2.3cm]{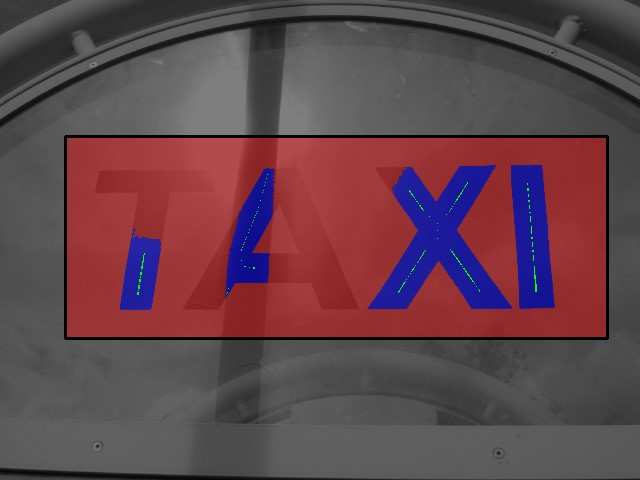} \\
\small(c) &\small (d) \\
\includegraphics[height=2.3cm]{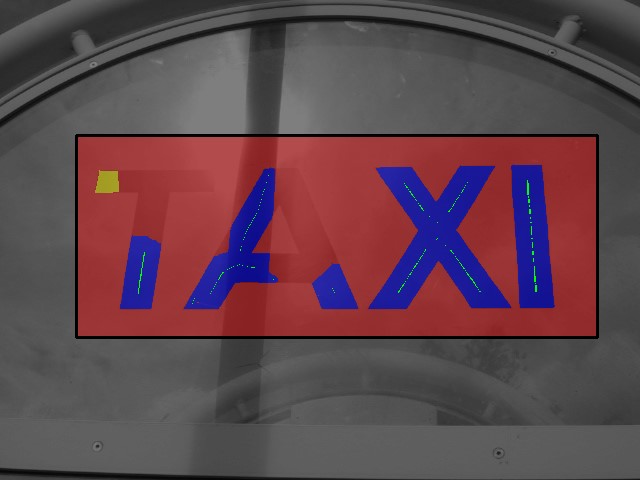} & \includegraphics[height=2.3cm]{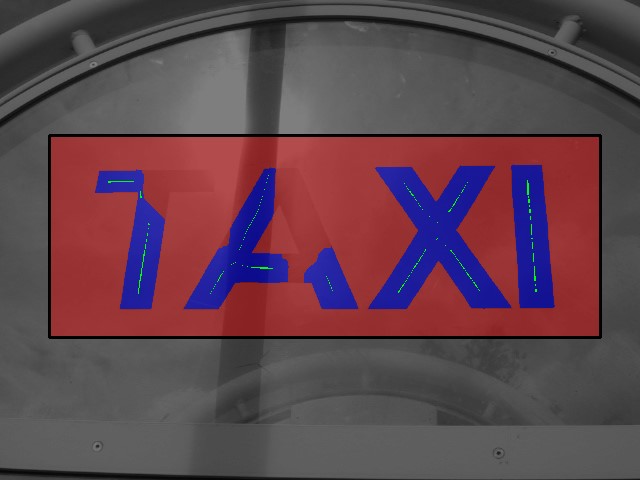} \\
\small(e) &\small (f) \\
\includegraphics[height=2.3cm]{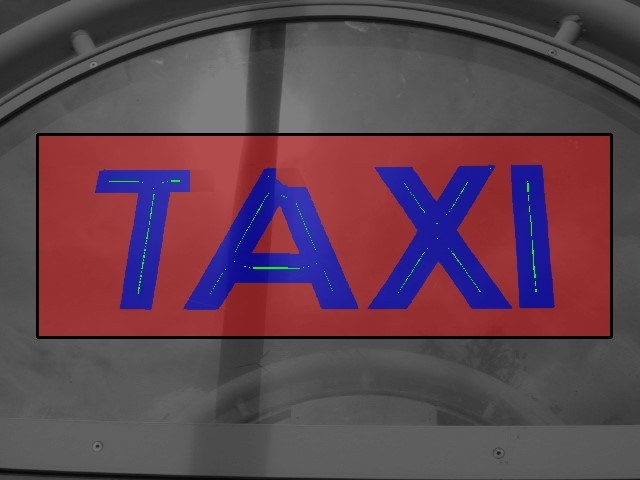} & \begin{tabular}{c}
\includegraphics[height=1.3cm, trim=0 130 0 70]{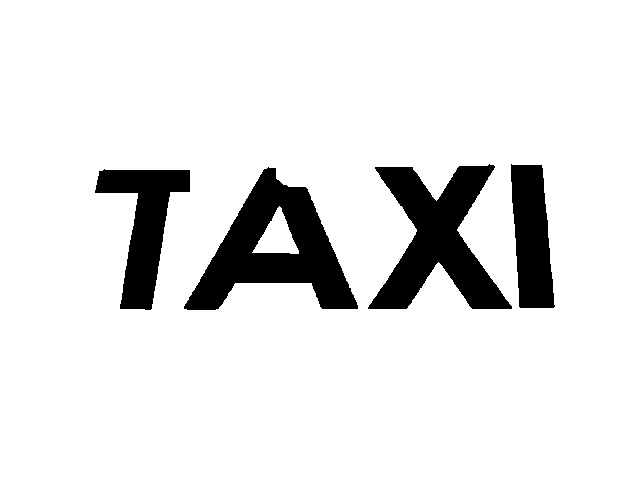} \\ \texttt{\textbf{TAXI}} \end{tabular} \\
\small(g) &\small (h) \\
\end{tabular}
\caption{The method pipeline. Source image (a). Initial MSER detection and classification (b) - character MSERs denoted green, multi-character MSERs blue and background MSERs denoted red. Text lines formation (c) - bottom line estimate in red. Local text refinement for the first text line - initialization (d), first iteration (e), second iteration (f), the last iteration (g), \emph{definitive foreground} pixels in green, \emph{probable foreground} pixels in blue, \emph{background} pixels in red, ignored pixels in yellow. Final segmentation and text recognition (h)}
\label{fig:pipeline}
\vspace{-15pt}
\end{figure}

Methods based on connected components~\cite{MicrosoftCVPR,Neumann-ACCV10,Yao-CVPR2012,Neumann-CVPR2012, Shi2013107, Neumann-ICDAR2013} find individual characters as connected components of a certain local property (color, intensity, stroke-width, etc.), so that the complexity is unaffected by the text parameters as characters of all scales and orientations can be detected in one pass. Moreover, the connected component representation provides a segmentation which can be exploited in a standard OCR stage. The main drawback is the assumption that a character is a single connected component, which is brittle - a change in a single image pixel introduced by noise can cause an unproportional change in the connected component size, shape or other properties. The methods are also incapable of detecting characters which consist of several connected components or where text is present as characters joint together.

In the proposed method, we generalize the region-based approach by detecting arbitrary fragments and groups of characters alongside characters themselves in a single stage. As previously suggested~\cite{MicrosoftCVPR}, we exploit the observation that text consists of strokes and we propose a unified approach to effectively detect and further segment regions which are formed of strokes, regardless whether they represent a part of a character, a whole character or a group of characters joint together, thus dropping the common assumption of a one to one correspondence between a character and its connected component representation.

\begin{figure}
\centering
\begin{tabular}{VV}
\includegraphics[trim=230 525 300 230, height=2cm]{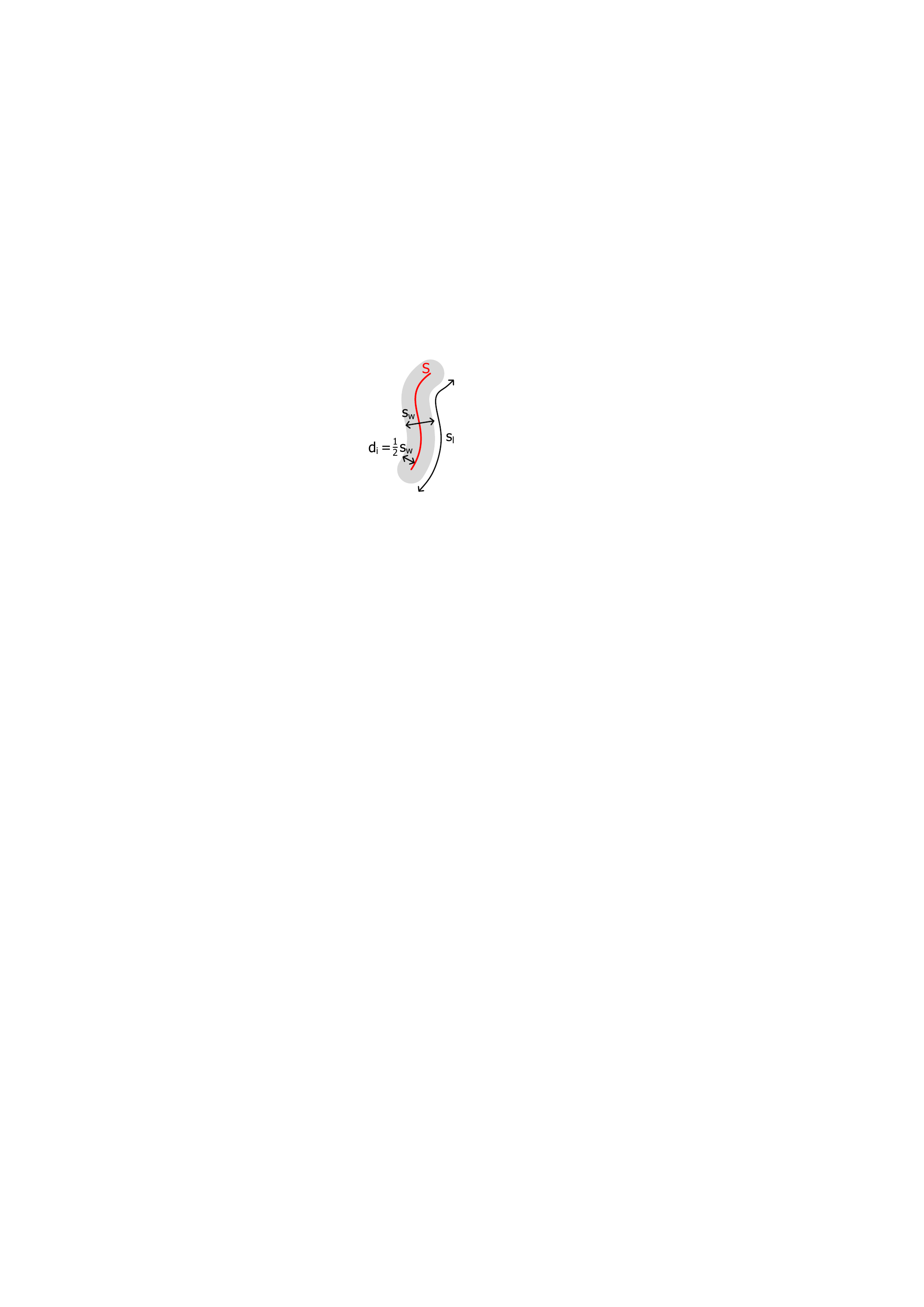} &
$A = s_w*s_l \doteq 2\sum\limits_{ i\in \mathbf{S}} d_i$
\end{tabular}
\caption{Area $A$ of an ideal stroke is a product of the stroke width $s_w$ and the length of the stroke $s_l$. This is approximated by summing double the distances $d_i$ of \emph{Stroke Support Pixels (SSPs)} along the stroke axis $\mathbf{S}$}
\label{fig:strokearea}
\vspace{-5pt}
\end{figure}

In the initial stage, candidate regions are effectively detected as MSERs~\cite{Matas-MSER} with the ``strokeness'' property and grouped into initial text line hypotheses, where each text line hypothesis is then individually segmented or rejected using an iterative and more robust segmentation approach, which is capable of segmenting characters that cannot be obtained by thresholding (and therefore neither as MSERs). In order to estimate the ``strokeness'' of a region we introduce a novel feature based on \emph{Stroke Support Pixels (SSPs)} which exploits the observation that one can draw any character by taking a brush with a diameter of the stroke width and drawing through certain points of the character (see Figure~\ref{fig:strokearea}) - we refer to such points as \emph{stroke support pixels (SSPs)}. The SSPs have the property that they are in the middle of a character stroke, which we refer to as the \emph{stroke axis}, the distance to the region boundary is half of the stroke width, but unlike skeletons they do not necessary form a connected graph.


Since the area (i.e. the number of pixels) of an ideal stroke is the product of the stroke width and the length of the stroke, the ``strokeness'' is estimated by the \emph{stroke area ratio} feature $\varsigma$ which compares the actual area of a region with the ideal stroke area calculated from the SSPs. The feature estimates the proportion of region pixels which are part of a character stroke and therefore it allows to efficiently differentiate between text regions (regardless of how many characters they represent) and the background clutter. The feature is efficiently computed from a region distance map, it is invariant to scaling and rotation and it is more robust to noise than methods which aim to estimate a single stroke width value~\cite{MicrosoftCVPR} as small pixel changes do not cause unproportional changes to the estimate. At last but not least, the SSPs are also exploited in the subsequent supervised segmentation stage to build a more accurate text color model, as by definition the SSPs are placed inside the character where the character color varies the least.

The rest of the paper is structured as follows: In Section \ref{sec:previousWork}, an overview of previously published methods is presented, in the Section \ref{sec:proposedMethod} the proposed method is introduced and in Section \ref{sec:experiments}, the experimental evaluation is given. The paper is concluded in the Section \ref{sec:conclusion}.

\vspace{-5pt}
\section{Previous Work}

\label{sec:previousWork}
Numerous methods which focus solely on text localization in real-world images have been published. The ``sliding-window'' based methods~\cite{Lee-ICDAR2011} use a window which is moved over the image and the presence of text is estimated on the basis of local image features. The majority of recently published methods for text localization however uses the connected component approach~\cite{Neumann-CVPR2012, Neumann-ICDAR2013, MicrosoftCVPR,Yao-CVPR2012,Mishra-CVPR2012,kang2013orientation}. The methods differ in their approach to individual character detection, which could be based on edge detection, character energy calculation or extremal region detection, but they all share the assumption that a single character is detected as a single connected component. The winning method in text localization of Yin et al.~\cite{Yin-TPAMI2013} at the latest ICDAR 2013 Robust Reading competition~\cite{ICDAR2013} also falls into this category as individual characters are detected as Maximally Stable Extremal Regions (MSERs)~\cite{Matas-MSER}.

Other methods focus only on text recognition, where the text is manually localized by a human annotator. The text is recognized on various levels, ranging from characters~\cite{Iwamura2013} to the whole word level~\cite{photoocr,yao2013strokelets,leeregion}. The winning method~\cite{photoocr} was able to correctly recognize $82.8\%$ of the manually cropped-words in the latest ICDAR Robust Reading competition~\cite{ICDAR2013}. Although the methods for cropped-word recognition give an upper-bound on currently achievable text recognition performance, they in fact assume there exists a text localization method with a $100\%$ accuracy, which currently is far from being true. Moreover, since the text was localized by a human, it is not clear that such text localization is even possible without the recognition, because the human annotator could have used the actual content of the text to create the annotation for localization.

For an exhaustive survey of text localization and recognition methods refer to the ICDAR Robust Reading competition results~\cite{ICDAR2013}.


\vspace{-10pt}
\section{The Proposed Method}
\label{sec:proposedMethod}
\subsection{Initial Candidates Detection}
\label{sec:initialCandidateDetection}

\begin{figure}
\centering
\includegraphics[width=0.75\columnwidth, trim=140 275 70 450]{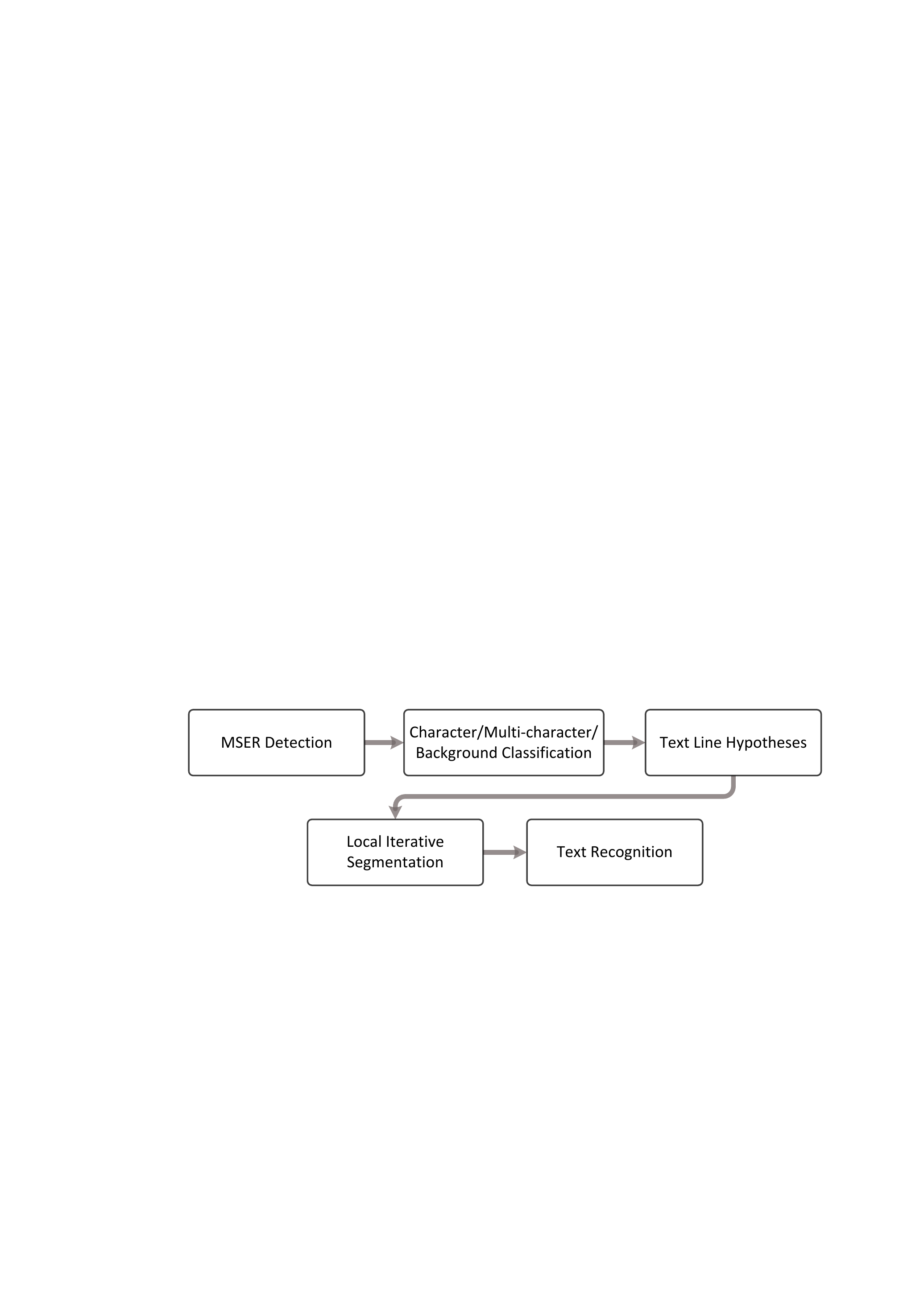}
\caption{Overview of the method. Initial text hypotheses efficiently generated by a MSER detector are further refined using a local text model, unique to each text line}
\label{fig:methodoverview}
\vspace{-15pt}
\end{figure}

\begin{figure}
\centering
\setlength{\tabcolsep}{1pt}
\begin{tabular}{cccccc}
\includegraphics[height=1.4cm]{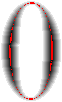} & \includegraphics[height=1.4cm]{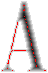} & \includegraphics[height=1.5cm]{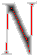} & \includegraphics[height=1.3cm]{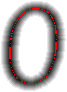} & \includegraphics[height=1.4cm]{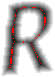} & \includegraphics[height=1.4cm]{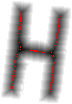}\\
\scriptsize $0.85$ & \scriptsize $0.75$ & \scriptsize $0.93$ \scriptsize & \scriptsize $0.95$ &\scriptsize  $0.74$ & \scriptsize $0.71$ \\
\multicolumn{3}{c}{\includegraphics[width=.40\columnwidth]{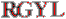}} & \multicolumn{3}{c}{\includegraphics[width=.40\columnwidth]{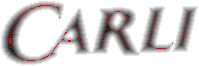}}
\\
\multicolumn{3}{c}{\scriptsize $0.82$} & \multicolumn{3}{c}{\scriptsize $0.71$} 
\\
\includegraphics[height=1.3cm]{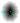} & \includegraphics[height=1.3cm]{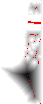} & \multicolumn{2}{c}{\includegraphics[height=1.4cm]{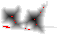}} & \multicolumn{2}{c}{\includegraphics[height=1.3cm]{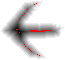}}\\
\scriptsize $0.17$ & \scriptsize $0.42$ & \multicolumn{2}{c}{\scriptsize $0.34$} & \multicolumn{2}{c}{\scriptsize $0.54$} \\
\end{tabular}
\caption{Examples of stroke area ratio $\varsigma$ values for character (top row), multi-character (middle row) and background (bottom row) connected components. Distance map denoted by pixel intensity, \emph{Stroke Support Pixels (SSPs)} denoted red}
\label{fig:distanceMap}
\vspace{-15pt}
\end{figure}

In the initial stage, candidate regions are detected as MSERs~\cite{Matas-MSER}. The MSER detector is often exploited in the literature~\cite{Yin-TPAMI2013,Neumann-ACCV10} to effectively detect individual characters, however this assumption may not always hold - there are many instances where individual characters cannot be detected as MSERs because only a portion of a character is a MSER (see Figure~\ref{fig:pipeline}b) or a single MSER corresponds to multiple characters or even whole words (see Figure~\ref{fig:distanceMap},~middle~row).

In the proposed method, we significantly relax the assumption of individual characters being detected as MSERs (or even Extremal Regions~\cite{Neumann-CVPR2012, Neumann-ICDAR2013}) by considering the MSER detector as an efficient first stage in order to generate initial text hypothesis, with no assumptions what level of text (i.e. part of characters, characters or words) individual MSERs represent. In other words, the proposed method assumes that at least a small portion of the text in the image triggers the MSER detector to generate an initial hypothesis, but it does require that all characters are detected as MSERs, as the individual characters are detected at a later stage using a local text model.

In order to build initial text hypotheses, all MSERs in an image (detected in the intensity and hue channels) are first classified into $3$ distinct classes: The \emph{character} class represents a single character (or a significant portion of it), the \emph{multi-characters} class represents an arbitrary group of characters joint together as a single component (e.g. a portion of a word, a whole word or even several words) and the \emph{background} class represents all non-textual content (e.g. background textures). The MSERs classified as \emph{characters} and \emph{multi-characters} are used to initialize a local text model (see Section~\ref{sec:localTextModel}), whilst the MSERs classified as \emph{background} are discarded.

For each region, the following features are calculated: stroke area ratio $\varsigma = \frac{A_s}{A}$, aspect ratio $\frac{w}{h}$, compactness $\frac{\sqrt{A}}{P}$, convex hull area ratio $\frac{A}{A_c}$ and holes area ratio $\frac{A_h}{A_c}$,
where $w$ and $h$ denote width and height of the region's bounding box, $A$ denotes the region area (i.e. number of pixels), $P$ denotes the length of the perimeter, $A_c$ denotes the convex hull area, $A_h$ denotes the total area of region holes and $A_s$ denotes the \emph{character strokes area}.

In order to estimate the character strokes area, a distance transform map is calculated for the region binary mask and only pixels corresponding to local distance maxima are considered (see Figure~\ref{fig:distanceMap}) - we refer to these pixels as \emph{Stroke Support Pixels (SSPs)}, because the pixels determine the position of a latent character stroke axis.

\begin{figure}
\setlength{\tabcolsep}{1pt}
\centering
\begin{tabular}{cc}
\includegraphics[trim=205 600 307 100, height=2.8cm]{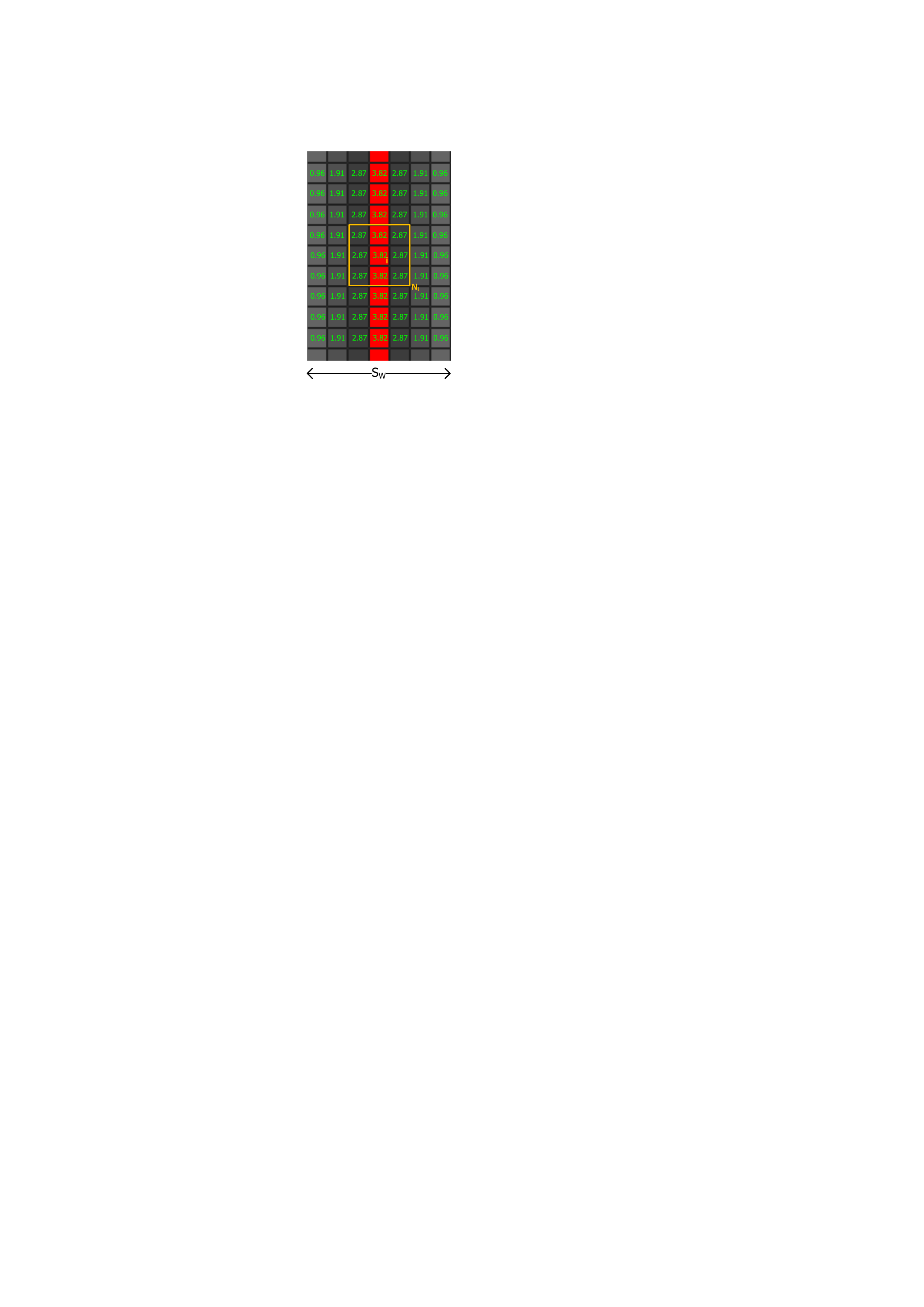} &
\includegraphics[trim=200 600 290 100, height=2.8cm]{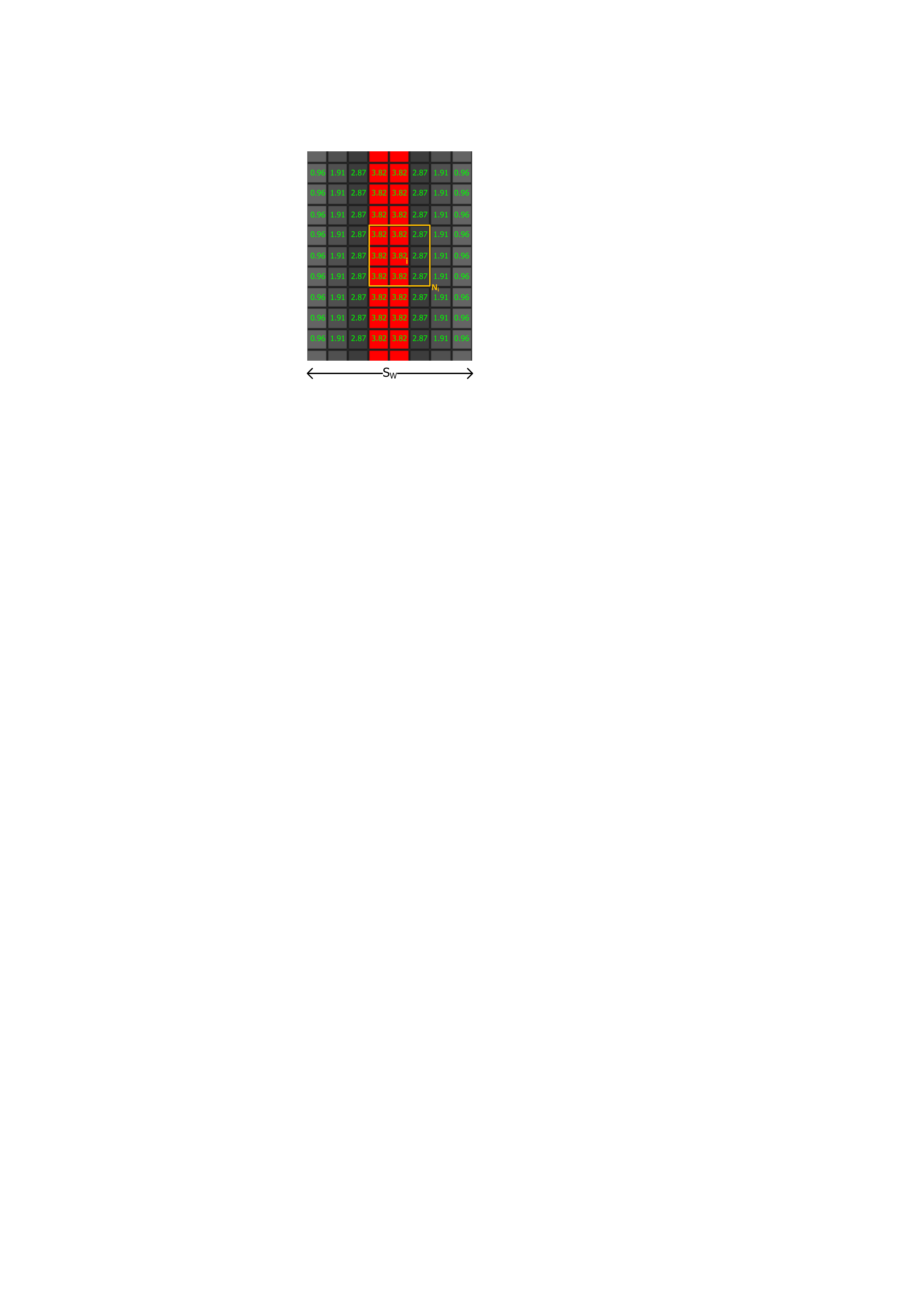} \\
\scriptsize
 {$\!\begin{aligned}
               A_s &= 2*9*\frac{3}{3}*3.82 = 68.76 \\
              \varsigma &= \frac{A_s}{A} = \frac{68.76}{70} = \textbf{0.98} \\
\end{aligned}$} &
\scriptsize
 {$\!\begin{aligned}
               A_s &= 2*18*\frac{3}{6}*3.82 = 68.76 \\
               \varsigma &= \frac{A_s}{A} = \frac{68.76}{80} = \textbf{0.86} \\
\end{aligned}$} \vspace{3pt} \\
\small (a) & \small (b) \\

\includegraphics[trim=200 510 195 90, width=.40\columnwidth]{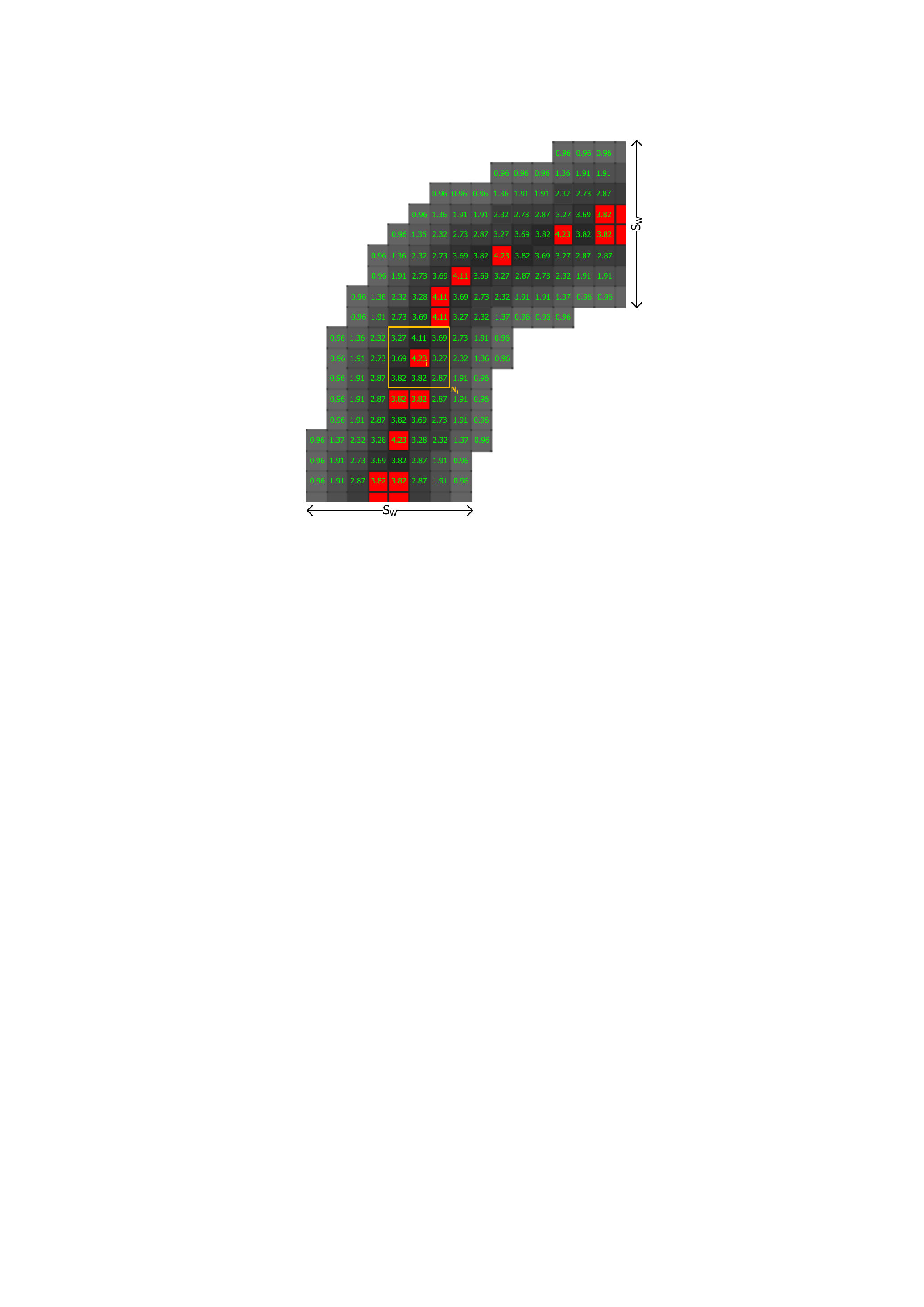} &
\includegraphics[trim=200 510 200 90, width=.40\columnwidth]{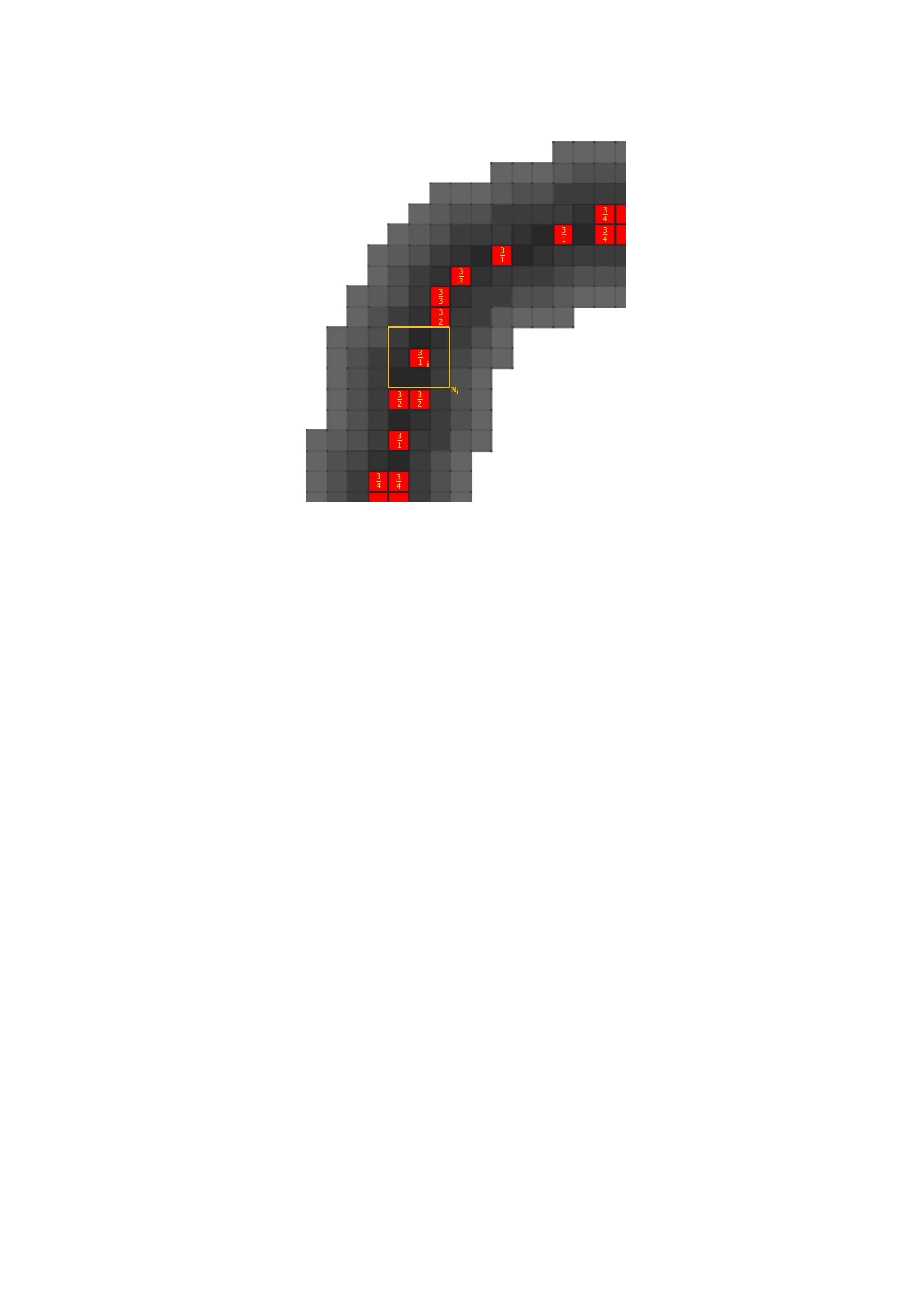} \\
\multicolumn{2}{c}{
\scriptsize
 {$\!\begin{aligned}
               A_s &= 2*(\frac{3}{4}*3.82+\frac{3}{4}*3.82+\frac{3}{1}*4.23+\dots+\frac{3}{4}*3.82) = 180.24\\
               \varsigma &= \frac{A_s}{A} = \frac{180.24}{187} = \textbf{0.96} \\

\end{aligned}$} \vspace{3pt}}  \\
\small (c) & \small (d) \\
\end{tabular}
\caption{Stroke area ratio $\varsigma$ calulation for a straight stroke of an odd (a) and even (b) width and for a curved stroke - distance map $d_i$ (c) and Stroke Support Pixel weights $w_i$ (d). Stroke Support Pixels (SSPs) denoted red}
\label{fig:strokes}
\vspace{-15pt}
\end{figure}

In order to estimate the area of the character strokes $\bar{A_s}$, one could simply sum the distances associated with the SSPs
\begin{equation}
\bar{A_s} = 2\sum_{ i\in \mathbf{S}} d_i
\end{equation}
where $\mathbf{S}$ are the SSPs and $d_i$ is the distance of the pixel $i$ to the boundary.

Such an estimate is correct for an straight stroke of an odd width, however it becomes inaccurate for strokes of an even width (because there are two support pixels for a unitary stroke length) or when the support pixels are not connected to each other as a result of noise at the region boundary or changing stroke width (see Figure ~\ref{fig:strokes}). We therefore propose to compensate the estimate by introducing weights $w_i$, which ensure normalization to a unitary stroke length by counting the number of pixels in a $3\times3$ neighborhood of each SSP
\begin{equation}
A_s = 2\sum_{ i\in \mathbf{S}} w_id_i \quad w_i=\frac{3}{|\mathcal{N}_i|}
\end{equation}
where $|\mathcal{N}_i|$ denotes the number of SSPs within the $3\times3$ neighborhood of the pixel of $i$ (including the pixel $i$ itself). The numerator value is given by the observation that for a straight stroke, there are $3$ support pixels in the $3\times3$ neighborhood (see Figure~\ref{fig:strokes}a).

To generate the training data, all MSERs from the ICDAR 2013 Training Set~\cite{ICDAR2013} dataset images were labeled using the ground truth segmentation masks - if the MSER overlaps sufficiently (more than $70\%$ of pixels) with a ground truth character segmentation it is labeled as a \emph{character}, if it overlaps with multiple ground truth character segmentations it is labeled as a \emph{multi-character} and if it does not overlap with any segmentation it is labeled as \emph{background}. MSERs which do not fall into any of the above categories were not used in the training. Using the aforementioned procedure, a dataset of $121{,}000$ background MSERs, $14{,}000$ character MSERs and $1{,}200$ multi-character MSERs was obtained. A random subset of $20{,}000$ samples was then used to train an SVM classifier~\cite{SVM} with a RBF~\cite{RBF} kernel using a \mbox{one-against-all strategy}, where each class was assigned a weight inversely proportional to its ratio in the training dataset in order to deal with the unbalanced number of samples for each class.

\vspace{-5pt}
\subsection{Text Line Hypotheses}
\label{sec:textLineHypotheses}

Given the initial set of text hypothesis in the form of detected character and multi-character regions, the proposed method proceeds to build a local text model. The model is inferred for each \emph{text line} individually, where we consider a \emph{text line} as a sequence of characters which can be fitted by a line and which has the same typographic and appearance properties.

The character and multi-character regions are first clustered into initial text line hypotheses using an efficient exhaustive search strategy adapted from~\cite{Neumann-ICDAR11}, where each neighboring character triplet and each multi-character region is assigned a bottom line estimate (see Figure~\ref{fig:pipeline}c), which serves as a distance measure for a standard agglomerative clustering approach. In order to enforce that one region is present only in one text line, initial text lines are simply grouped into clusters based on presence of identical regions (two text lines are a member of the same cluster if they have at least one region in common) and then in each cluster only the longest text line is kept; this can be viewed as a voting process, where in each cluster text lines vote for the text direction and the longest text line wins.

\subsection{Local Iterative Segmentation}
\label{sec:localTextModel}

Each text line hypothesis is further refined using a local text model, individual for each text line. We formulate the problem of finding the local text model as a energy minimization task in a standard graph cut framework by adapting the iterative segmentation approach of GrabCut~\cite{rother2004grabcut} by dynamically changing the processed image area based on current segementation in each iteration.

Let us recall that in the graph cuts framework the objective is a minimization of a the Gibbs energy
\begin{equation}
E(\alpha, \theta, z) = U(\alpha,\theta,z) + V(\alpha, z)
\end{equation}
where $U(\alpha,\theta,z)$ is the data term, $V(\alpha, z)$ is the smoothness term, $\alpha$ is the vector of labels for each pixel, $\theta$ represents the image color distributions for background and foreground and $z$ is the image.

Following~\cite{rother2004grabcut}, the data term is a Gaussian Mixture Model (one GMM for foreground and one for background) and the smoothness term is the based on the Euclidean distance in the RGB color space. Each pixel within the text line bounding-box is then labeled as \emph{definitive foreground} \texttt{DF}, \emph{probable foreground} \texttt{PF} or \emph{background} \texttt{B} in the following iterative process (see Figure~\ref{fig:pipeline}d-g):
\begin{enumerate}
  \item Initialize all pixels belonging to a character or a multi-character region as \texttt{PF}, others as \texttt{B}
  \item Calculate a new text line bounding-box by encapsulating all \texttt{PF} pixels and expand it by $\gamma_h$ and $\gamma_v$ pixels in the horizontal resp. vertical direction
  \item Find SSPs amongst \texttt{PF} pixels and mark them as\texttt{DF}
  \item Learn GMM parameters, using the \texttt{DF} pixels to train the foreground model and \texttt{B} pixels to train the background model
  \item Create edges from the source to the \texttt{DF} and \texttt{PF} pixels, and from the \texttt{B} pixels to the sink
  \item Estimate the segmentation by finding the minimal cut - mark pixels in the source subgraph as \texttt{PF}, pixels in the sink subgraph as \texttt{B}
  \item Repeat from Step 2, until convergence
\end{enumerate}

The value $\gamma_h$ is set to the average region width in the text line and the $\gamma_v$ is one third of the text line bounding-box height. 

The final segmentation of the text line is a obtained by taking the connected components of the \texttt{PF} pixels. If all pixels in the text line bounding-box converged to the same label (e.g. all are labeled as \texttt{PF}), the whole text line is discarded as it most likely represents a false positive. Pixels with the \texttt{PF} label which do not fit the bottom line estimate (see Figure~\ref{fig:pipeline}e) or which are located at the boundary of the text line bounding-box are ignored in the GMM estimation as they typically represent interpunction or fragments of characters in neighbouring text lines.

\begin{figure}
\centering
\setlength{\tabcolsep}{1pt}
\begin{tabular}{|c|c|}
\hline
\includegraphics[height=2cm]{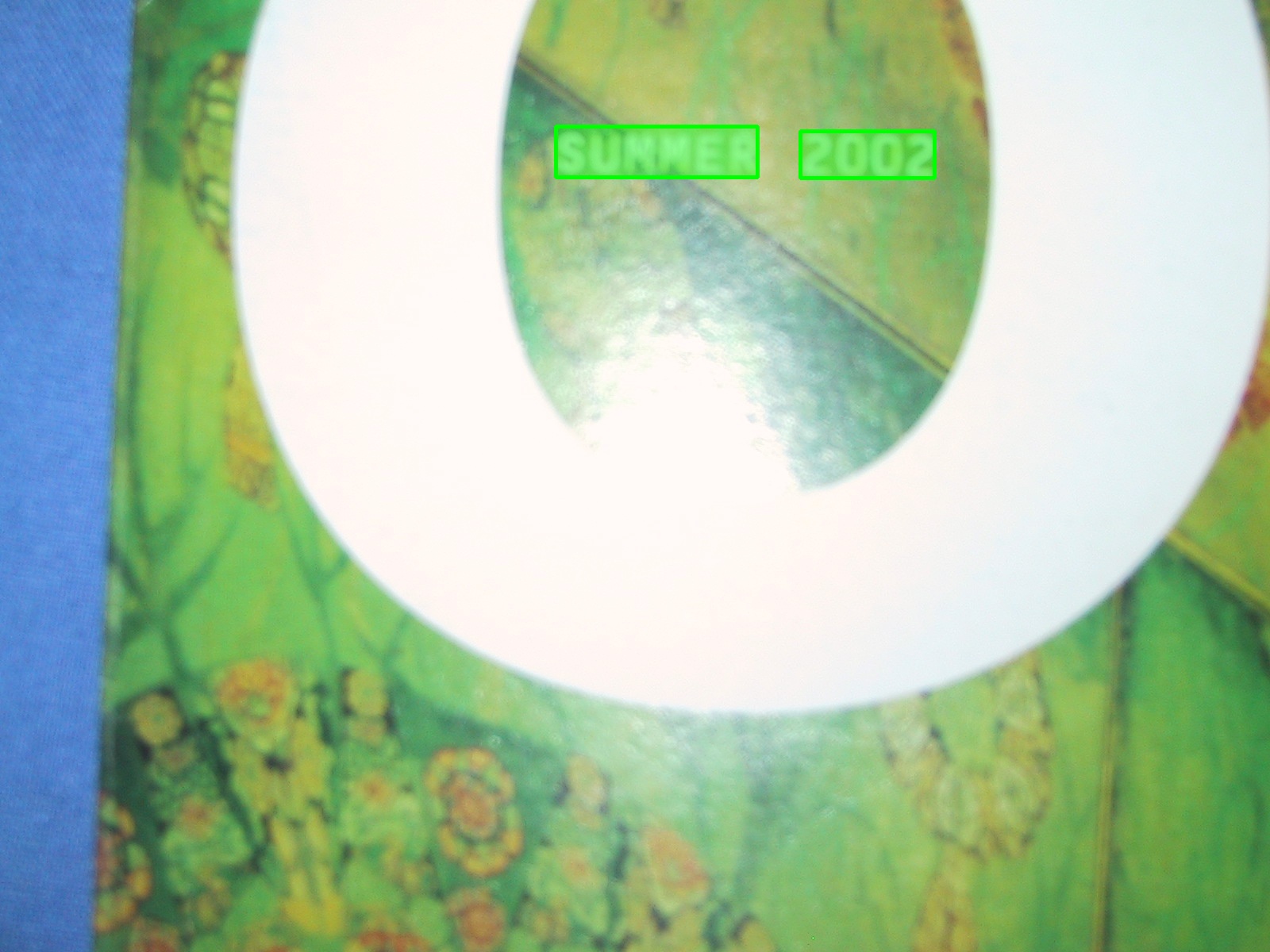}&
\includegraphics[height=2cm]{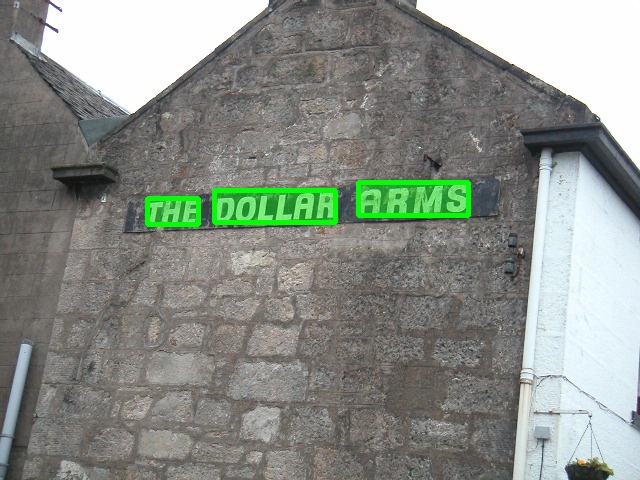} \\
\scriptsize\begin{tabular}{c}\texttt{SUMMER 2002}\end{tabular}&
\scriptsize\begin{tabular}{c}\texttt{THE DOLLAR ARMS}\end{tabular} \\ \hline
\includegraphics[height=2cm]{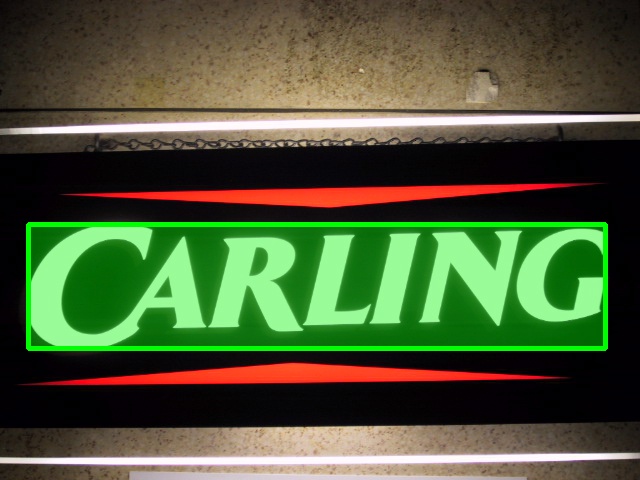}&
\includegraphics[height=2cm]{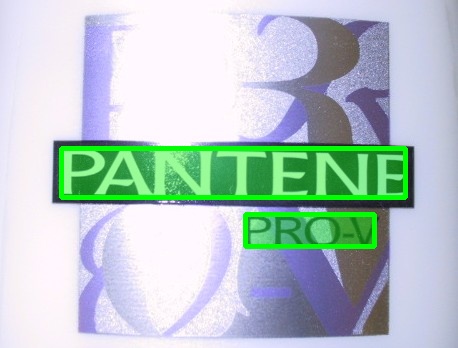} \\
\scriptsize\begin{tabular}{c}\texttt{CARLING}\end{tabular}&
\scriptsize\begin{tabular}{c}\texttt{PATENE PROV}\end{tabular} \\ \hline
\includegraphics[height=2cm]{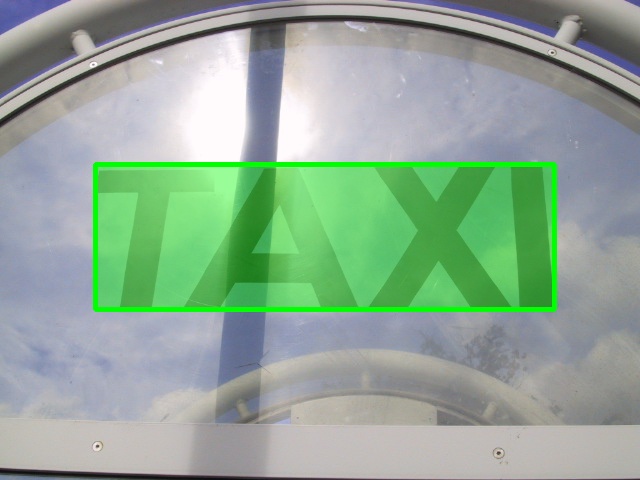}&
\includegraphics[height=2cm]{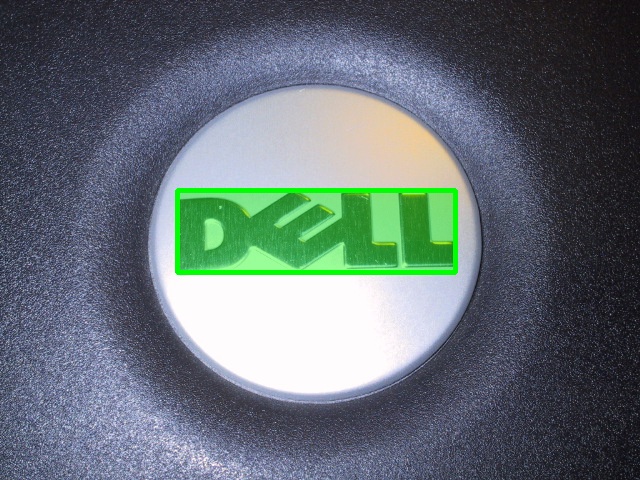} \\
\scriptsize\begin{tabular}{c}\texttt{TAXI}\end{tabular}&
\scriptsize\begin{tabular}{c}\texttt{D8LL}\end{tabular} \\ \hline
\includegraphics[height=2cm]{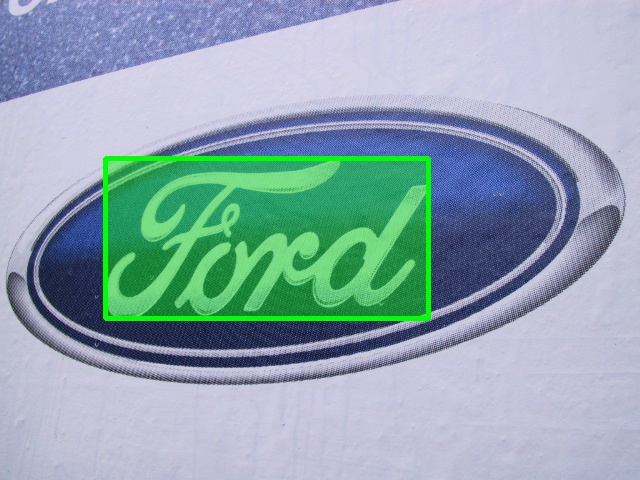}&
\includegraphics[height=2cm]{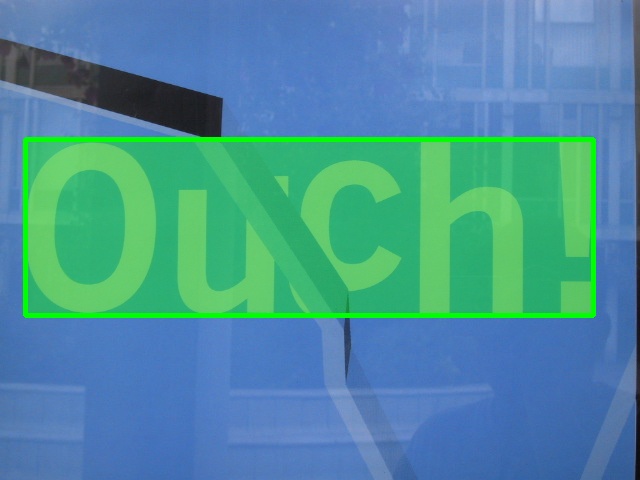} \\
\scriptsize\begin{tabular}{c}\texttt{Ym8}\end{tabular}&
\scriptsize\begin{tabular}{c}\texttt{ouchI}\end{tabular} \\ \hline
\end{tabular}
\caption{Text localization and recognition examples on the ICDAR 2013 dataset}
\label{fig:icdar2013output}
\vspace{-10pt}
\end{figure}


\vspace{-10pt}
\subsection{Text Recognition}
\label{sec:textRecognition}

Given the segmentations obtained in the previous stage, each connected component is assigned a Unicode label(s) by an OCR module, which is trained on synthetic data~\cite{Neumann-ACCV10}. Following the standard approach~\cite{tesseract}, the connected components with the aspect ratio above a predefined threshold are chopped to generate more region hypotheses in order to cater for joint characters. Each connected component with a label then represents a node in a direct acyclic graph, where the edges represent a ``is-a-predecessor'' relation. The final sequence of labels is then found as an optimal path in such a graph~\cite{Neumann-ICDAR2013}.

Because the graph is relatively small (when compared to \cite{Neumann-ICDAR2013}, where there are several segmentations for each character), second order language model features were added in order to improve recognition accuracy without any significant impact on memory complexity.

The whole pipeline runs independently over multiple scales for each image and in the final stage the detected words are aggregated into a single output, while eliminating overlapping words (which typically represent the same word detected in multiple scales) by keeping only the word whose corresponding path in the graph has the lowest cost.



\section{Experiments}
\label{sec:experiments}
The proposed method was evaluated using the ICDAR 2013 Robust Reading competition dataset~\cite{ICDAR2013}, which contains $1189$ words and $6393$ letters in $255$ images.

\begin{table}
\caption{Comparison with most recent results on the ICDAR 2013 dataset.}
\label{table:icdarLocalization}
\centering
\begin{tabular}{|c||cc|c||c|}
\hline
method & recall & precision & f & published \\ \hline
\textbf{proposed method} & \textbf{72.4} & \textbf{81.8} & \textbf{77.1} & \\
Yin et al.~\cite{Yin-TPAMI2013} & 68.3 & 86.3 & 76.2 & 2014 \\
TexStar (ICDAR'13 winner)~\cite{ICDAR2013} & 66.4 & 88.5 & 75.9 & 2013 \\
our previous method~\cite{Neumann-ICDAR2013} & 64.8 & 87.5 & 74.5 & 2013 \\
Kim (ICDAR'11 winner)~\cite{ICDAR2011} & 62.5 & 83.0 & 71.3 & N/A \\
\hline
\end{tabular}
\vspace{-10pt}
\end{table}

Using the ICDAR 2013 competition evaluation scheme~\cite{ICDAR2013}, the method achieves recall $72.4\%$, precision $81.8\%$ and f-measure $77.4\%$ in text localization (see Figure \ref{fig:icdar2013output} for sample outputs). The method achieves significantly better recall than the winner of ICDAR 2013 Robust Reading competition ($66\%$) and the method of Yin et al.~\cite{Yin-TPAMI2013} ($68\%$) and outperforms all the previous methods in the f-measure - see Table \ref{table:icdarLocalization}.

In end-to-end text recognition, the method correctly localized and recognized $543$ words ($45\%$), where a word is considered correctly recognized if all its characters match the ground truth (using case-sensitive comparison). On the other hand, the method ``hallucinated'' $79$ words in total which do not have any overlap with the ground truth.

The main reasons for method failure are character-like objects near the text (e.g. pictographs, arrows, etc.) and low-contrast characters which are not picked up in the initial stage. The average run time on a standard 2.7GHz PC is $800$ms per image.

\vspace{-10pt}
\section{Conclusion}
\label{sec:conclusion}
An end-to-end real-time text localization and recognition method was presented. The method detects initial text hypothesis in a single pass by an efficient region-based method and subsequently refines the text hypothesis using a more robust local text model, which deviates from the common assumption of region-based methods that all characters are detected as connected components.

Additionally, a novel feature based on Stroke Support Pixels (SSPs) is introduced. The feature is based on an observation, that one can draw any character by taking a brush with a diameter of the stroke width and drawing through certain points of the character. The feature is efficiently computed from a region distance map, it is invariant to scaling and rotations and allows to efficiently detect text regions regardless of what portion of text they capture.

On the standard ICDAR 2013 dataset~\cite{ICDAR2013}, the method achieves state-of-the-art results in text localization (f-measure $77.6\%$) and improves previously published results for end-to-end text recognition, with the average run time of $800$ms per image.

Future work includes dealing with current limitations of the method, such as the inability to detect single- or two-letter words if they are not part of a longer text line and the assumption of a straight base-line in the text line hypothesis stage.

\vspace{-10pt}
\bibliographystyle{IEEEtranS}
\bibliography{icdar2015}

\end{document}